\documentclass[conference]{IEEEtran}
\IEEEoverridecommandlockouts

\usepackage{hyperref}
\usepackage[usenames,dvipsnames]{xcolor}

\definecolor{bblue}{HTML}{4F81BD}
\definecolor{rred}{HTML}{C0504D}
\definecolor{ao}{rgb}{0.0, 0.5, 0.0}

\usepackage{booktabs}
\usepackage{pgfplots}
\usepackage{tikz}
\usepackage{multirow}

\def\footurl#1{\footnote{\url{#1}}}

\usepackage[utf8]{inputenc}
\usepackage[T1]{fontenc}
\usepackage[upright]{fourier}
\usepackage{pgfplotstable} 
\usetikzlibrary{arrows,fit}
\usepackage{cite}
\usepackage{amsmath,amssymb,amsfonts}
\usepackage{algorithmic}
\usepackage{textcomp}
\usepackage{fancyhdr}
\def\BibTeX{{\rm B\kern-.05em{\sc i\kern-.025em b}\kern-.08em
    T\kern-.1667em\lower.7ex\hbox{E}\kern-.125emX}}
\usepackage{cleveref}
\usepackage{graphicx}
\usepackage{subcaption}    
\graphicspath{ {./images/} }
\usepackage{array} 
\usepackage{float}

\begin{document}
\title{Analyzing COVID-19 Vaccination Sentiments in Nigerian Cyberspace: Insights from a Manually Annotated Twitter Dataset\\
\thanks{\rule[3pt]{\columnwidth}{0.4pt}

Funding received from HausaNLP and Arewa Data Science Academy.}
}

\author {\IEEEauthorblockN{Ibrahim Said Ahmad}
\IEEEauthorblockA{\textit{Northeastern University}\\
Boston, USA \\
isab7070@gmail.com}
\and
\IEEEauthorblockN{Lukman Jibril Aliyu}
\IEEEauthorblockA{\textit{HausaNLP}\\
Kano, Nigeria \\
lukman.j.aliyu@gmail.com}
\and
\IEEEauthorblockN{Auwal Abubakar Khalid}
\IEEEauthorblockA{\textit{Bayero University}\\
Kano, Nigeria \\
aka2000078.mcs@buk.edu.ng}
\and
\IEEEauthorblockN{Saminu Muhammad Aliyu}
\IEEEauthorblockA{\textit{Bayero University}\\
Kano, Nigeria \\
smaliyu.cs@buk.edu.ng}

\and 
\and
\IEEEauthorblockN{Shamsuddeen Hassan Muhammad}
\IEEEauthorblockA{\textit{University of Porto}\\
Porto, Portugal \\
shamsuddeen2004@gmail.com}
\and
\IEEEauthorblockA{Idris Abdulmumin}
\IEEEauthorblockA{\textit{Ahmadu Bello University}\\
Zaria, Nigeria \\
iabdulmumin@abu.edu.ng}
\and
\IEEEauthorblockN{Bala Mairiga Abduljalil}
\IEEEauthorblockA{\textit{University Of Maiduguri}\\
Borno, Nigeria \\
ballaabduljalil@gmail.com}
\and
\IEEEauthorblockN{Bello Shehu Bello}
\IEEEauthorblockA{\textit{Bayero University}\\
Kano, Nigeria \\
bsbello.cs@buk.edu.ng}
\and
\IEEEauthorblockN{Amina Imam Abubakar}
\IEEEauthorblockA{\textit{University of Abuja}\\
Abuja, Nigeria \\
aminaImamabubakar@gmail.com}
}

\maketitle
\thispagestyle{fancy}

\begin{abstract}
Numerous successes have been achieved in combating the COVID-19 pandemic, initially using various precautionary measures like lockdowns, social distancing, and the use of face masks. More recently, 
 various vaccinations have been developed to aid in the prevention or reduction of the severity of the COVID-19 infection.
Despite the effectiveness of the precautionary measures and the vaccines, there are several controversies that are massively shared on social media platforms like Twitter. In this paper, we explore the use of state-of-the-art transformer-based language models to study people’s acceptance of vaccines in Nigeria. We developed a novel dataset by crawling multi-lingual tweets using relevant hashtags and keywords. Our analysis and visualizations revealed that most tweets expressed neutral sentiments about COVID-19 vaccines, with some individuals expressing positive views, and there was no strong preference for specific vaccine types, although Moderna received slightly more positive sentiment. We also found out that fine-tuning a pre-trained LLM with an appropriate dataset can yield competitive results, even if the LLM was not initially pre-trained on the specific language of that dataset.

\end{abstract}


\section{Introduction}
Over the last decade, social media has provided abundant data that researchers often utilized to derive insights into people's sentiments. As such, several research areas have emerged relying on social media data, including sentiment analysis, emotion detection, stance detection and social network analysis \cite{cui2023survey}. In this research, we explored the use of tweets in Nigerian cyberspace for the prediction of people's opinions towards the COVID-19 vaccine. This approach is practical because it allows for real-time feedback, is cost-effective, provides a large reach, offers unfiltered opinions, and enables trend analysis.

The COVID-19 pandemic is a global health crisis caused by the spread of the novel coronavirus SARS-CoV-2 \cite{al2022ensemble}. It has had far-reaching and profound impacts on societies, economies, healthcare systems, and daily life worldwide. The pandemic had a significant impact on public health and society at large, necessitating the rapid development and distribution of vaccines to combat the virus. While the availability of vaccines is crucial, so too is public sentiment towards them \cite{li2022dynamic,blanco2022optimism, yin2022sentiment}. In Nigeria, social media platforms such as Twitter have played a crucial role in shaping public discourse around COVID-19 \cite{obi2020social,adamu2021framing}. As such, the discussions on Twitter provide crucial data that can be used to study COVID-19 vaccination in Nigeria. Therefore, we crawl tweets in the Nigerian cyberspace using the discontinued Twitter Academic API\footurl{https://www.cnn.com/2023/04/05/tech/academic-researchers-blast-Twitter-paywall/index.html} from 1st November 2020 to 30th June 2022 using COVID-19 vaccine keywords. A total of 5200 tweets were collected. We preprocessed the tweets by removing URLs, tweets with less than three words, and tweets not in Latin characters, so the tweets were reduced to 4320. Therefore 4,320 tweets were manually annotated by three (3) independent annotators. \par

We conducted exploratory data analysis to derive insights from the dataset. The results of the analysis reveal that the majority of Nigerians either believe that vaccines are secure and life-saving or hold a neutral stance on the vaccines, while a smaller fraction of Nigerians have expressed concerns about the vaccine's safety. Additionally, we found that the majority of the people in Nigeria hold a neutral stance on COVID-19 brands, notwithstanding, a small percentage of the people still hold a negative stance towards the brands. \par

We conducted experiments to demonstrate that a strong performance can be achieved on COVID-19 acceptance prediction using the developed dataset by both classical machine learning models and state-of-the-art pre-trained large language models. The experimental results reveal that the support vector machine achieved the best performance amongst the classical machine learning models while \emph{xlm-roberta base} achieved the best performance amongst the large language models. \par

\noindent
\textbf{Contributions:} The main contributions of this paper are:
\begin{enumerate}
    \item We developed a manually annotated COVID-19 vaccination acceptance dataset in Nigerian cyberspace. 
    \item We analyzed the data and derived novel insights from the data using multiple exploratory data analysis techniques like topic modelling. 
    \item We benchmark the dataset on COVID-19 vaccination sentiment analysis using both classical machine learning techniques and large language models. 
\end{enumerate}

The remainder of this paper is organized as follows: Section 2 summarises the related works. Section 3 provides a detailed description of the research methodology. Section 4 presents the results and discussion. Finally, Section 5 concludes the paper with recommendations and future works.  
\par

\section{Related work} 
\paragraph {\textbf{COVID-19 Vaccine Sentiment Analysis:}}
People's opinion towards the COVID-19 vaccine has been a subject of study in the research community and sentiment analysis is among the most widely used approaches for this task \cite{aygun2021aspect, melton2021public,rahmanti2022social}. A study by \cite{baldha2021covid} collected and analysed tweets on COVID-19 to understand and extract the most discussed topics. They found "Vaccination programme" to be the widely discussed topic while \cite{sarirete2021bibliometric} conducted a similar research and found "vaccine hesitancy", "vaccine acceptance" and "vaccine safety" as the most discussed topic related to the COVID-19 vaccine. The study in \cite{li2022dynamic} used social media data to analyse people's opinions towards the acceptance of COVID-19 vaccines in the US. The vaccine acceptance index (VAI) was used to determine how well people accept these vaccines. The result of their experiment shows a positive increase in the acceptance of the COVID-19 vaccines in the last quarter of 2021. This result contradicts the findings of \cite{fridman2021covid} that says most people are not willing to administer the vaccine.
A related study by \cite{shah2023sentiment} analysed tweets related to people's acceptance of the Pfizer vaccine and found that most people have negative perceptions of the vaccine. 

\paragraph {\textbf{Fine-tuning Large Language Models:}} Pre-trained language models are trained on a large amount of data which makes them very popular in NLP tasks. Researchers now prefer fine-tuning instead of training a model from scratch, especially for small datasets \cite{schoene2023example}. Fine-tuning entails changing the weights of these pre-trained models on a new task to achieve better performance \cite{jiang2019smart}. Keshavarzi et al. \cite{keshavarzi2020artificial} fine-tuned BERT \cite{devlin2018bert} to investigate misinformation on Covid-19 vaccine from tweets. Similarly, \cite{cheon2023analysis} used BERT to predict COVID-19 vaccine's side effects using data from social media. The study in \cite{wei2021finetuned} fine-tuned RoBERTa \cite{liu2019roberta} to classify sentiments in COVID-19 vaccine tweets.
\section{Research Methodology}


\subsection{Data collection and Annotation}
We collected tweets from Twitter (now X) using the Twitter Academic API from 1st November 2020 to 30th June 2022. We collected tweets keywords relating to COVID-19 vaccination using keywords like vaccin*, moderna, pfizer, biontech, gamaleya, janssen, astrazeneca, astrazeneca, covishield, sinopharm, jab, microchips, infertility, tracking, magnet. The collection was restricted to only Tweets in Nigeria using location ID in the query. A total of 5200 tweets were collected. After removing URLS, tweets with less than three words, and tweets not in Latin characters, the tweets were reduced to 4320. \par  

The tweets were then annotated by three (3) independent annotators using an annotation guideline. The classes for the annotation are:
\begin{enumerate}[left=0pt, itemsep=0.5em]
   \item Positive: If the user’s tweet implies a positive attitude or opinion towards the vaccine (directly or indirectly). For example, the tweets contain positive words or some terms that imply a positive attitude towards COVID-19 vaccine like save lives, protect, appreciative, thankful, excited, or optimistic.
   \item Negative: If the user’s tweet implies a negative attitude or opinion towards the vaccine. For example, the tweets contain negative words or some terms that imply a negative attitude towards COVID-19 vaccine like kill, harmful, tracker, chip, critical, angry, disappointed in, pessimistic, expressing sarcasm about, or mocking COVID-19 vaccine.
   \item Neutral: If the user’s tweet does not imply any opinion. 
   \item Indeterminate: If the tweet is unintelligible. 
\end{enumerate}

After the annotation, \emph{majority vote} was used to assign the final class of tweets in the following manner: 
    \begin{enumerate}
        \item If all the annotators agree on the same class, then that class is automatically accepted. 
        \item If two annotators agree on either positive or negative and the last annotator classifies the tweet as neutral, then the tweet is assigned the majority class. 
        \item If two annotators classify a tweet as positive and the third annotator classifies it as negative, then the tweet is adjudicated by a new annotator. This is also the case if two annotators classify a tweet as negative and the third annotator classifies the tweet as positive.
        \item If all the annotators label the tweet differently, the tweet is discarded. 
    \end{enumerate}
At the end of the annotation, 4,320 tweets were annotated. \Cref{tab:annotated_tweets} gives an example of tweets in the dataset and their annotated classes.

\begin{table}[t]
\begin{center}
\renewcommand{\arraystretch}{1.5} 
\caption{\label{tab:annotated_tweets} Example of tweets and their annotated classes }
\addtolength{\tabcolsep}{2pt}
\begin{tabular}{p{0.5cm} p{5cm} p{1.1cm}} 
\hline \hline
\multicolumn{1}{c}{SN} & \multicolumn{1}{c}{Tweet} & \multicolumn{1}{c}{Class} \\ \hline

1 & i took my COVID vaccine jab today monday at the flag off of the fct vaccination campaign i believe it is safe praise god & Positive  \\
2 & taking a vaccine that was developed under a year goes against every damn thing we know about drug discovery and vaccine creation refer back to the vaccines are not effective if they were why are vaccinated people being asked to wear masks & Negative \\
3 & COVID vaccine is safe and effective ive gotten mine go and get yours & Positve \\
4 & so who didnt approve the madagascarn herb but the astrazeneca vaccine they approved is now giving issues all over the world oh sorry like this was what naija imported hmmmm & Negative \\
5 & gov makinde ssg others vaccinated as oyo govt kicks off COVID vaccination & Neutral \\

\hline
\end{tabular}
\end{center}
\end{table}    

\subsection{Exploratory Data Analysis (EDA)}
Our approach is focused on analyzing public sentiment on Twitter towards COVID-19 vaccines. The initial step we explored is Exploratory Data Analysis (EDA), aimed at gaining valuable insights from the tweets. EDA entails the examination and interpretation of information and relevant data to draw conclusions regarding people's opinions. Therefore, we visualized the most common words used in COVID-19 discussions using the word clouds. Additionally, we used bar chat to compare the values of positive, negative and neutral tweets in our dataset. We also visualize the people's sentiments towards the most common COVID-19 vaccines.



\subsection{Models}
For the prediction of people's sentiment towards COVID-19 vaccine, we utilize two categories of models. First, we model our prediction based on classical machine learning algorithms. We used classical machine learning algorithms because our dataset is unlikely to yield reasonable results with deep learning models as they require a large training dataset. Additionally, classical machine learning algorithms sometimes yield competitive results compared to deep learning models despite being less computationally expensive. The machine learning algorithms used are Support Vector Machines (SVM), Logistic Regression, Naive Bayes, K-Nearest Neighbours (KNN) and Decision Trees. \par

Secondly, we model our prediction by fine-tuning large language models. Fine-tuning is an effective method of utilizing pre-trained large language models for downstream NLP tasks using available datasets. The large language models we fine-tuned are Bert, Afro-XMLR, and XLM-roBERTa.

\section {Results and Discussion}

\subsection{Findings from Exploratory Data Analysis}
\begin{figure*}[ht]
  \centering
  \begin{subfigure}[b]{0.48\linewidth}
    \includegraphics[width=\linewidth]{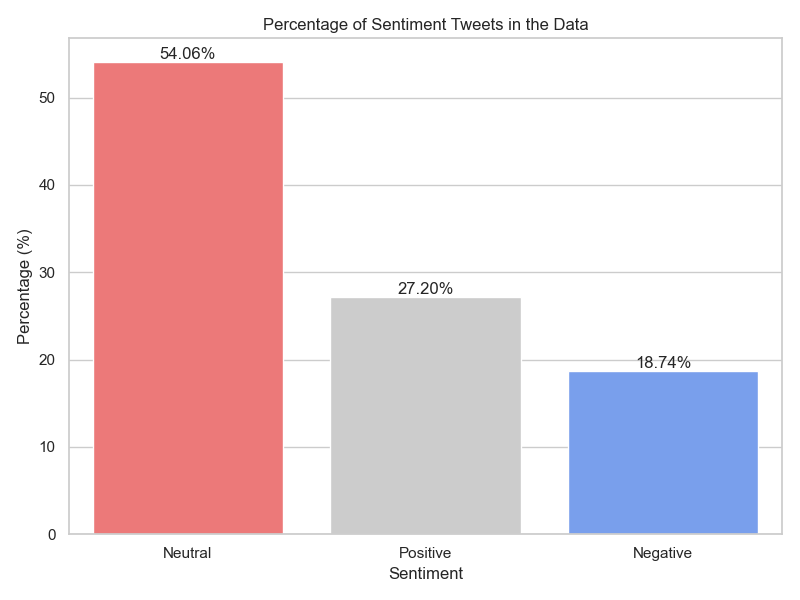}
    \caption{Percentage of positive, negative, and neutral tweets}
    \label{FIG:sentiment_distribution}
  \end{subfigure}
  \hfill
  \begin{subfigure}[b]{0.48\linewidth}
    \includegraphics[width=\linewidth]{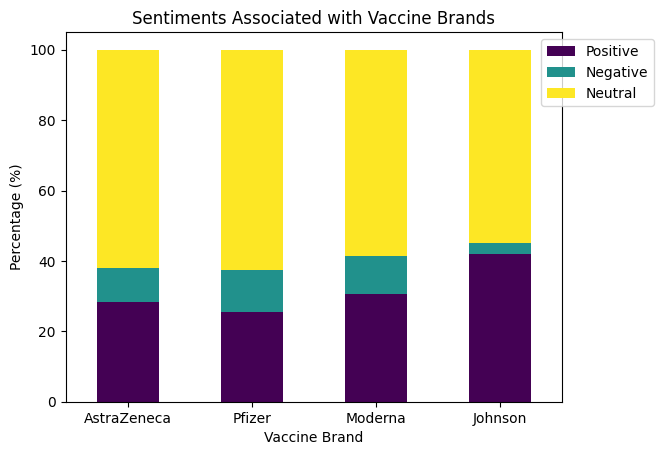}
    \caption{Sentiments of vaccine common brands}
    \label{FIG:vaccine_brand_percentage}
  \end{subfigure}
  
  \begin{subfigure}[b]{0.48\linewidth}
    \includegraphics[width=\linewidth]{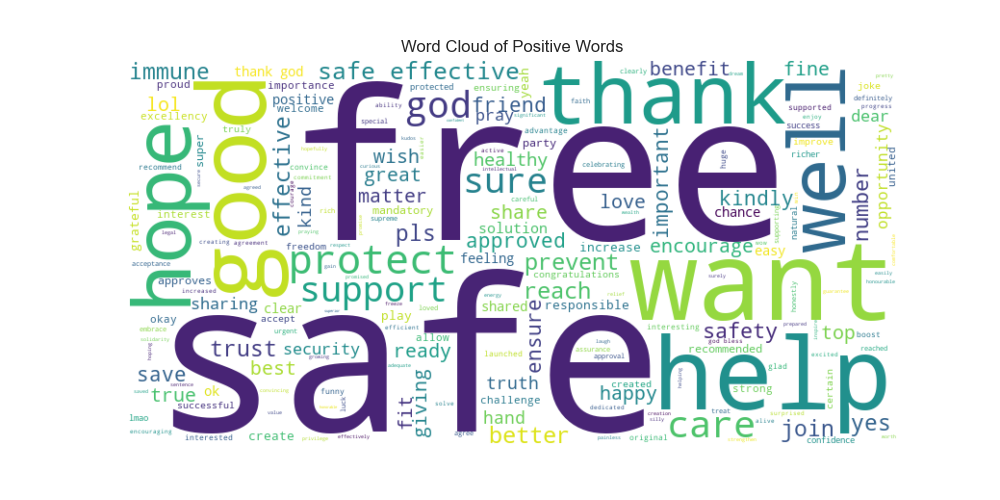}
    \caption{Word cloud for positive tweets }
    \label{FIG:positive_word_cloud}
  \end{subfigure}
  \hfill
  \begin{subfigure}[b]{0.48\linewidth}
    \includegraphics[width=\linewidth]{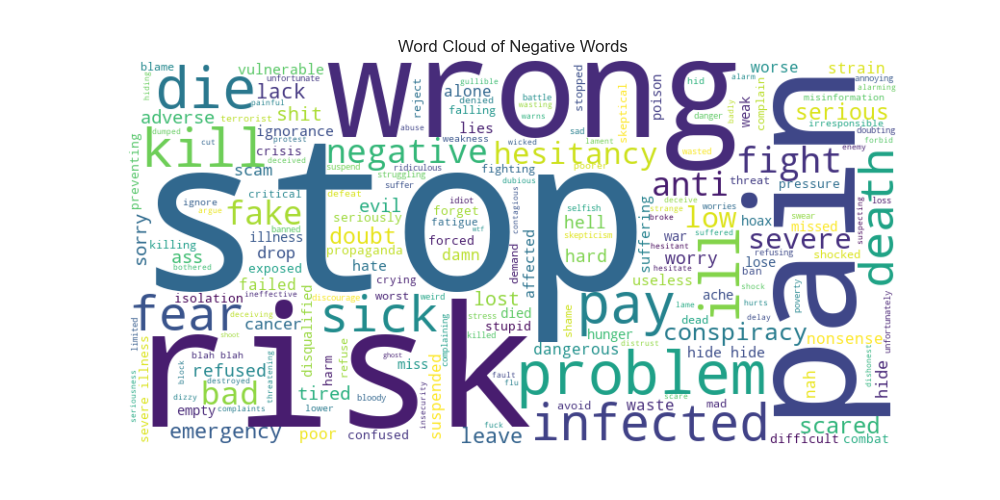}
    \caption{Word cloud for negative tweets}
    \label{FIG:negative_word_cloud}
  \end{subfigure}

  \caption{Exploratory data analysis}
  \label{Fig:exploratory_data-analysis}
\end{figure*}

\cref{Fig:exploratory_data-analysis} illustrates the findings from the EDA. \Cref{FIG:sentiment_distribution} shows the sentiment distribution of people towards COVID-19 vaccines. This information helps gauge whether Nigerians have embraced or rejected the safety of the vaccines. The results show that more than 27\% of the tweets carry positive sentiment towards the vaccines, while 18\%  of the tweets carry negative sentiment. Therefore, we can deduce that based on Twitter discussions, the majority of Nigerians either believe that vaccines are secure and life-saving or hold a neutral stance on the vaccines, while a smaller fraction of Nigerians have expressed concerns about the vaccine's safety.

In \cref{FIG:vaccine_brand_percentage}, we visualize people's stance towards  the most common COVID-19 vaccines, AstraZeneca, Pfizer, Moderna, and Johnson \& Johnson. It can be observed that the majority of the people hold a neutral stance on the brands, notwithstanding, a small percentage of the people still hold a negative stance towards the brands. Johnson \& Johnson vaccine has the best stance with less than 5\% of the tweets having a negative sentiment. 

\Cref{FIG:positive_word_cloud,,FIG:negative_word_cloud} show the most common words people use in conveying their opinion towards COVID-19 vaccine. It can be observed that when individuals hold a positive sentiment towards vaccines, they employ words such as \emph{safe, free, help, and protect}. These words suggest that vaccination has been successful for a segment of the population who view the vaccine as a significant lifesaving measure during challenging times. Conversely, some individuals hold opposing views, citing instances where people died despite being vaccinated. In such cases where individuals hold negative sentiments, they use words like \emph{risk, kill, pain, problem, and wrong} to convey their concerns. \par

\subsection{Classical Machine Learning Models}

\Cref{tab:ML} shows the performance of classical machine learning algorithms on COVID-19 vaccine sentiment classification. SVM and Logistic Regression achieved the best performance performance with an F1-score of $65\%$. However, SVM outperformed the decision tree in terms of accuracy. Naive Bayes perform poorly with an F1-score of $45\%$ and accuracy of $57\%$. Similarly, the Decision tree achieved poor results with F1-score of $53\%$ and an accuracy of $53\%$. 

\begin{table}[t]
\begin{center}
\caption{\label{tab:ML} Performance of Classical Machine Learning Models }
\begin{tabular}{lllll} 
\hline \hline
Model &  Accuracy & Precision & Recall & F1- Score\\ \hline
\textbf{SVM} & 0.66 & 0.65 & 0.66 & \textbf{0.65}  \\
Logistic Regression & 0.61 & 0.63 & 0.61 & 0.65 \\
Naive Bayes & 0.57 & 0.65 & 0.57 & 0.45  \\
K Nearest Neighbours & 0.62 & 0.61 & 0.62 & 0.61 \\
Decision Trees & 0.53 & 0.53 & 0.53 & 0.53 \\
\hline
\end{tabular}
\end{center}
\end{table}

\subsection{Fine-tuned Large Language Models}

\Cref{tab:LLM} shows the evaluation results of fine-tuned LLMs. The models exhibited better performance, which aligns with expectations. XLM-Roberta and Afro-XLMR achieved the best with and F1-score of $71\%$ and $69\%$ respectively, and an accuracy of $71\%$. These two algorithms have been trained on African languages and as such were expected to attain superior performance due to the prevalence of Nigerian English code-mixed with other Nigerian languages in the dataset. Notwithstanding, the bert model also performed competitively with F1-score and accuracy of $70\%$. The overall effectiveness of LLMs can be attributed to their extensive training using diverse natural language texts. From our analysis of the results, it is evident that fine-tuning a pre-trained LLM with an appropriate dataset can yield competitive results, even if the LLM was not initially pre-trained on the specific language of that dataset.\par

\begin{table}[t]
\begin{center}
\caption{\label{tab:LLM} Performance of Fine-tuned Large Language Models }
\addtolength{\tabcolsep}{2pt}
\begin{tabular}{lllll} 
\hline \hline
Model &  Accuracy & Precision & Recall & F1- Score\\ \hline
Bert base   & 0.70 & 0.70 & 0.70 & 0.70  \\
Afro-xmlr base  & 0.71 & 0.70 & 0.71 & 0.69 \\
\textbf{xlm-roberta base} & 0.71 & 0.71 & 0.71 & \textbf{0.71} \\
\hline
\end{tabular}
\end{center}
\end{table}

\section{Conclusion and Recommendation}
In this work, we developed and publicly released a manually annotated tweet dataset for COVID-19 vaccination exploration in Nigerian cyberspace. The dataset consists of 4320 annotated tweets. In addition, we employed multiple exploratory data analysis techniques to derive insights from the dataset. We found that most tweets indicate people generally have a neutral sentiment towards the COVID-19 vaccine. Notwithstanding, a good number of people also hold positive sentiments towards COVID-19 vaccines. We also found out that the majority of the people in Nigeria have no significant preference for the common vaccine types, even though the Moderna vaccine has slightly more positive sentiments compared to the other vaccines.    

In Addition to the dataset, we also investigate the efficiency of fine-tuning LLMs for COVID-19 vaccine acceptance prediction in contrast to classical machine learning algorithms using a small dataset. We found out that all the machine learning algorithms performed below the fine-tuned LLMs. This finding can be attributed to the fundamental differences between these two model categories. While LLMs, predominantly based on transformer architectures have been trained to grasp contextual information, classical machine learning models primarily focus on predicting labels without considering contextual nuances. Among the classical machine learning algorithms, SVM outperformed the other algorithms in terms of macro average F1-score. This superiority can be elucidated by the algorithm's ability to capture non-linear relationships within the data points.

Future work should dig further to fine-tune the hyperparameters and preprocessing steps to obtain the best model performance. Moreover, the limitation of our study is the relatively small dataset, comprising fewer than four thousand samples, coupled with label imbalance. These factors may constrain the performance of the fine-tuned models.

\section*{Ethics Statement}
This work does not raise any ethical concerns.

\section*{Acknowledgements}
This work was made possible by the mentorship program of the HausaNLP research group.

\bibliographystyle{IEEEtran}
\bibliography{conference}

\end{document}